\documentclass[11pt,a4paper]{article}
\usepackage{acl2020}
\usepackage{times}
\usepackage{latexsym}
\usepackage{graphicx}
\usepackage{supertabular}
\usepackage{longtable}
\usepackage{booktabs}
\usepackage{amsfonts,amsmath,balance}

\usepackage{microtype}
\usepackage{amsthm,stmaryrd,color,graphicx,multirow,paralist,pifont,subfigure,url,xspace,threeparttable}
\usepackage{amssymb}

\aclfinalcopy 

\newcommand{\textred}[1]{\textcolor{red}{\textbf{#1}}}

\newcommand{\textgreen}[1]{\textcolor{green}{\textbf{#1}}}
\newcommand{\cmarkb}{\textgreen{\ding{52}}}
\newcommand{\xmark}{\textred{\ding{55}}}

\newtheorem{definition}{Definition}

\title{A Probabilistic Model with Commonsense Constraints for Pattern-based Temporal Fact Extraction}

\author{Yang Zhou$^1$, Tong Zhao, Meng Jiang \\
Department of Computer Science and Engineering \\
University of Notre Dame, Notre Dame, IN 46556 \\
\texttt{\{yzhou24, tzhao2, mjiang2\}@nd.edu}}

\begin{document}
\maketitle

\begin{abstract}
Textual patterns (e.g., \texttt{Country}'s president \texttt{Person}) are specified and/or generated for extracting factual information from unstructured data. Pattern-based information extraction methods have been recognized for their efficiency and transferability. However, not every pattern is reliable: A major challenge is to derive the most complete and accurate facts from diverse and sometimes conflicting extractions. In this work, we propose a probabilistic graphical model which formulates fact extraction in a generative process. It automatically infers true facts and pattern reliability without any supervision. It has two novel designs specially for temporal facts: (1) it models pattern reliability on two types of time signals, including temporal tag in text and text generation time; (2) it models commonsense constraints as observable variables. Experimental results demonstrate that our model significantly outperforms existing methods on extracting true temporal facts from news data.
\end{abstract}

\footnotetext[1]{This work was done when the first author was a visiting undergraduate student at Notre Dame.}

\section{Introduction}
\label{sec:introduction}

Temporal fact extraction is to extract (entity, value, time)-factual tuples from text data (e.g., news, tweets) for specific attributes. It acts as one of the fundamental tasks in knowledge base construction, knowledge graph population, and question answering. For example, if we were interested in \textit{country's president}, the entity would be of type \texttt{Location.Country}, the value would be of type \texttt{Person}, and the time would be a valid year in the person's presidential term.
Thanks to name entity recognition (NER) and typing systems \cite{del2015finet}, pattern-based information extraction methods generate patterns consisted of entity types \cite{jiang2017metapad,qli2018truepie,reimers2016temporal}. They are widely used for good transferability across domains and datasets, unsupervised manner requiring no or very few annotations, and high efficiency. The typed patterns give only the association between entity and value. Two types of time signals can be attached to the pairs, forming temporal triples: One is temporal tag in text, e.g., the year tag next to the entity/value mentions in the sentence; the other is text generation time, i.e., the year the text document was posted. For example, given two sentences:

\noindent 1) ``\textit{... The former \underline{French} [\texttt{Country}: France] president \underline{Jacques Chirac} [\texttt{Person}], a self-styled affable rogue who was head of state from \underline{1995} [temporal tag] to 2007 ...}'' (posted on Sept. 26, \underline{2019} [text generation time])

\noindent 2) ``\textit{... \underline{Emmanuel Macron} [\texttt{Person}], now President of \underline{France} [\texttt{Country}], graduated from ENA in \underline{2004} [temporal tag] ...}'' (posted on Sept. 19, \underline{2019} [text generation time])

Pattern-based methods discover two patterns:
\begin{compactitem}
\item {P1:} former \texttt{Country} president \texttt{Person}
\item {P2:} \texttt{Person}, now president of \texttt{Country},
\end{compactitem}
Then the methods can extract the following tuples. We label $\cmarkb$ and $\xmark$ for correct tuples and incorrect ones, respectively: \\
\noindent $\cmarkb$ {\small (France, Jacques Chirac, 1995): {P1} and {temporal tag};} \\
\noindent $\xmark$ {\small (France, Jacques Chirac, 2019): {P1} and {text gen. time};} \\
\noindent $\xmark$ {\small (France, Emmanuel Macron, 2004): {P2} and {temporal tag};} \\
\noindent $\cmarkb$ {\small (France, Emmanuel Macron, 2019): {P2} and {text gen. time}.}

We have the following observations:
\begin{compactitem}
    \item \textbf{O1:} Not every pattern is reliable: the pattern ``\texttt{Person} visited \texttt{Country}'' is very likely to be unreliable. Not every pattern is unreliable: the pattern ``current \texttt{Country}'s president \texttt{Person}'' is very likely to be reliable. The above two pattern examples are somehow half and half. So, patterns have reliability.
    \item \textbf{O2:} For temporal fact extraction, different types of time signals might be either reliable or unreliable depending on the pattern. So, there is a dependency between pattern and type of time signal, in terms of reliability.
\end{compactitem}

Existing truth finding approaches assumed that a structured ``source-object-claim'' database was given and then estimated the reliability of source for inferring whether the claim was true or false \cite{yin2008truth,zhao2012probabilistic,zhao2012bayesian}. For example, a source could be a book seller, an object could be a book's author list, and a claim could be an author list that a seller gave for a book. One conclusion was that \textit{probabilistic graphical models} (PGM) \cite{zhao2012probabilistic,zhao2012bayesian} have advantages of estimating source reliability over the general data distributions, compared with bootstrapping algorithms \cite{yin2008truth,qli2018truepie,wang2019novel}. However, PGM-based truth finding models have not yet been developed for the task of \textit{information extraction}. Estimating the reliability of textual patterns is new (O1). Moreover, when we focus on temporal fact extraction, modeling the dependency between pattern and type of time signals is also new (O2).

In truth finding, it is critical to define conflicts. For the book seller's example, we assume that one book can have only one true author list; so if we knew one list was true, then any different list of the same book would be false. This originated from our commonsense. Fortunately, we have quite a few commonsense rules for temporal facts, i.e., specific attributes. On \textit{country's president}, we know that
\begin{compactitem}
    \item one president serves only one country;
    \item one country has only one president at a time;
    \item however, one country can have multiple presidents in the history (e.g., USA, France).
\end{compactitem}
For the attribute \textit{sports team's player}, we have commonsense rules:
\begin{compactitem}
    \item one player serves only one club at a time;
    \item however, one club has multiple players and one player can serve multiple clubs in his/her career.
\end{compactitem}
We generalize possible commonsense rules:
\begin{compactitem}
    \item {C1:} one value matches with only one entity;
    \item {C2:} one entity matches with only one value;
    \item {C3:} one value matches with only one entity at a time;
    \item {C4:} one entity matches with only one value at a time.
\end{compactitem}
So, we know that the attribute \textit{country's president} follows C1 and C4; and the attribute \textit{sports team's player} follows C3. 
The third challenge (O3) is the necessity of modeling the commonsense (e.g., C1--C4) for identifying conflicts, estimating pattern reliability, and finding true temporal facts.

To address the three challenges (O1--O3), we propose a novel \underline{P}robabilistic \underline{G}raphical \underline{M}odel with \underline{C}ommonsense \underline{C}onstraints (PGMCC), for finding true temporal facts from the results from pattern-based methods. The given input is the observed frequency of tuples extracted by a particular pattern and attached with a particular type of time signal. We model information source as a pair of pattern and type of time signal. We represent the source reliability as an unobserved variable. It becomes a generative process. We first generate a source. Next we generate a (entity, value, time)-tuple. Then we generate the frequency based on the source reliability and the tuple's trustworthiness (i.e., probability of being a truth). Moreover, we generate variables according to the commonsense rules if needed -- the variable counts the values/entities that can be matched to one entity/value with or without a time constraint (at one time) from the set of \textit{true} tuples. Given a huge number of patterns (i.e., 57,472) and tuples (i.e., 116,631) in our experiments, our proposed unsupervised learning model PGMCC can effectively estimate pattern reliability and find true temporal facts.

Our main contributions are:
\begin{compactitem}
\item {We introduce the idea of PGM-based truth finding to the task of pattern-based temporal fact extraction.}
\item {We propose a new unsupervised probabilistic model with observed constraints to model the reliability of textual patterns, the trustworthiness of temporal tuples, and the commonsense rules for certain types of facts.}
\item {Experimental results show that our model can improve AUC and F1 by more than 7\% over the state-of-the-art.}
\end{compactitem}

The rest of this paper is organized as follows. Section \ref{sec:problem} introduces the terminology and defines the problem. Section \ref{sec:approach} presents an overview as well as details of the proposed model. Experimental results can be found in Section \ref{sec:experiments}. Section \ref{sec:related} surveys the literature. Section \ref{sec:conclusions} concludes the paper.

\section{Terminology and Problem Definition}
\label{sec:problem}

\subsection{Terminology}

\begin{definition}[Temporal fact: (entity, value, time)-tuple)]
    Let $\mathcal{F} = \{f_1, f_2, f_3, \dots\}$
    be the set of temporal facts. Each fact $f$ is in the format of (entity, value, time). $\mathcal{F}$ was extracted by textual pattern-based methods.
\end{definition}

\begin{definition}[Pattern s] 
Let $\mathcal{P}^{(*)} = \{p^{(*)}_1, \dots\}$ be the set of pattern source, here $* \in \{post, tag\}$ stands for the type of time signal (i.e., ``text gen. time'' and ``temporal tag''). One pattern paired with different types of time signals will be treated as different pattern sources.
\end{definition}

\begin{definition}[Extraction] Let $\mathcal{E} = \{e_1, e_2, e_3, \dots\}$ be the set of extractions. Our generative model will take $\mathcal{E}$ as input. An extraction item $e$ is in the format of ($f$, $p^{(*)}$, $o$). Here $o$ stands for the observed frequency of fact tuples $f$ that were extracted by pattern ${p}^{(*)}$ in $\mathcal{E}$.
\end{definition}

\begin{definition}[Constraint]
Each commonsense rule (constraint) is represented as a variable. The variable is likely to be observed as 1. Examples:
\begin{compactitem}
\item one \texttt{value} matches with only one \texttt{entity}, denoted as $\mathcal{C}_{1v-1e}$ that counts the number of such entities.
\item one \texttt{entity} at one \texttt{time} matches with only one \texttt{value}, denoted as $\mathcal{C}_{1(e,t)-1v}$ that counts the number of values.
\end{compactitem}
\end{definition}

\subsection{Problem Definition}

Suppose the set of extractions $\mathcal{E}$ have been obtained by pattern-based methods from text data. We define the problem as follows:
\textbf{Given} a set of extractions $\mathcal{E}$, pattern sources $\mathcal{P}^{(*)}$, and the constraints $\mathcal{C}_{a}$ for attribute $a$, \textbf{infer} truth $\mathcal{T}$ for all temporal facts $\mathcal{F}$ contained in $\mathcal{E}$ and quality information for each pattern source $p^{(*)}$.

\section{Proposed Approach}
\label{sec:approach}

\begin{figure}[t]
\centering
\includegraphics[width=0.5\textwidth]{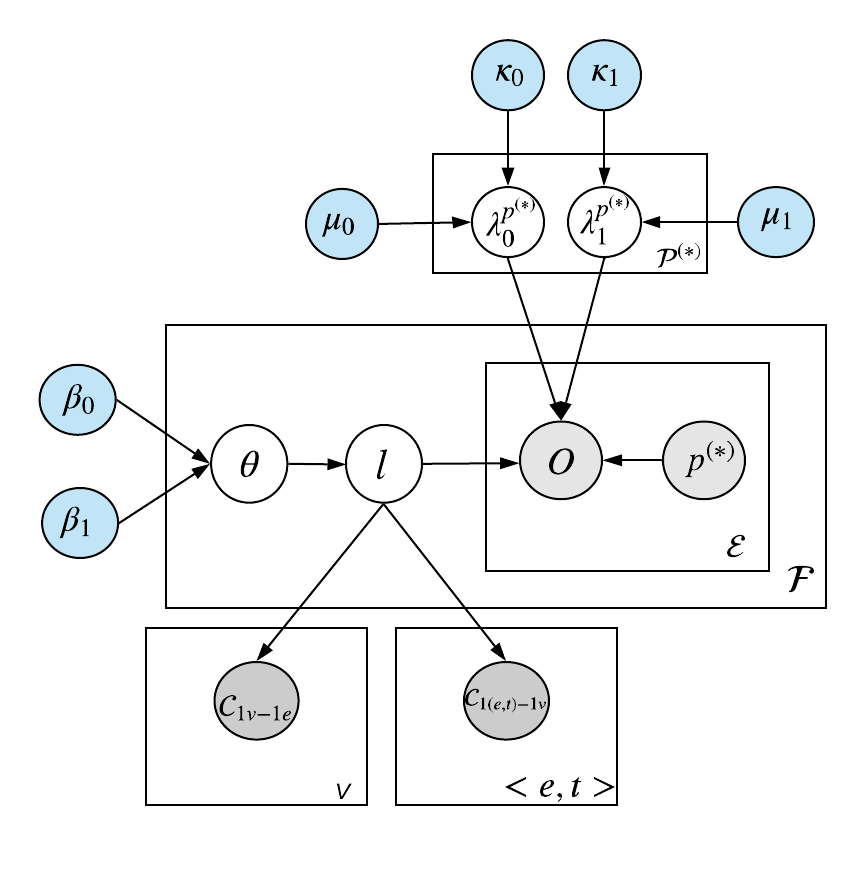}
\caption{Probabilistic Graphic Model with Commonsense Constraint \{$  \mathcal{C}_{1(e,t)-1v}$, $\mathcal{C}_{1(v)-1e}$ \}}
\label{pgm_fig}
\end{figure}

We mainly discussed the model detail of PGM with multiple Constraints $\mathcal {C}_{1(e,t)-1v}$  and  $\mathcal{C}_{1v-1e}$, since it's the most complicated scenario while modeling constraint.
The given input is the observed frequency of fact tuples extracted by a particular pattern and attached with a particular type of time signal. 
Figure~\ref{pgm_fig} gives the plate notation of our model. Each node represent a variable. Blue nodes indicate hyper-parameter. Gray nodes stand for observable variable. And white nodes stand for latent variables we want to infer.

\begin{table}[t]
\centering
\caption{Symbols and their descriptions.}
\begin{tabular}{|c|p{0.6\columnwidth}|}
\hline
\textbf{Symbol} & \textbf{Description} \\
\hline
$\theta_f$ & $[0,1]$, trustworthiness of temporal fact tuple $f$ \\
\hline
$l_f$ & Boolean: label of temporal fact $f$ \\
\hline
$o_e $& Integer: the observed frequency of fact $f_e$ extracted by pattern ${p}^{(*)}_e$ \\
\hline
$\lambda_0^{{p}^{(*)}},\lambda_1^{{p}^{(*)}} $ & Real numbers: reliability of pattern ${p}^{(*)}$ on giving false/true fact tuples \\
\hline
$\mathcal{C}_{1v-1e} $ & Real number: the number of entities given one value $v$\\
\hline
$\mathcal{C}_{1(e,t)-1v} $ & Real number: the sum of values given one entity $e$ and one time $t$\\
\hline \hline
\multicolumn{2}{|c|}{\textbf{Hyper-Parameter}}\\
\hline
$\mu_0, \mu_1 $ & Integers: prior counts of false/true tuples extracted by a textual pattern \\
\hline
$\kappa_0, \kappa_1 $ & Integers: prior sums of false/true tuples extracted by a textual pattern \\
\hline
$\beta_0, \beta_1 $ & Integers: prior counts of false/true tuples \\
\hline
\end{tabular}
\label{tab:notation}
\end{table}

\subsection{Generative Process}
 Our approach based on PGM is a generative process. We first generate a source. Next we generate a (entity, value, time)-tuple. Then we generate the frequency based on the source reliability and the tuple's trustworthiness. Moreover, we generate variables according to the commonsense constraints. The variables counts the values/entities that can be matched to one entity/value with or without a time constraint (at one time) from the set of true tuples.
The concrete meaning of each variable has been given in Table~\ref{tab:notation}. 

\vspace{0.05in}
\noindent \textbf{Temporal fact trustworthiness.} For each temporal fact $f \in \mathcal{F}$, we first draw $\theta_f$, i.e., the prior truth probability of fact $f$, from a \textit{Beta} distribution with hyper-parameter $\beta_0$ and  $\beta_1$:
\begin{equation}
\theta_f \sim Beta(\beta_0,\beta_1).
\end{equation}
$\beta_0$ and $\beta_1$ represent the prior distribution of fact reliability. In practice, if we have a strong prior knowledge about how likely all or certain temporal facts are true, we can model it with the corresponding hyper-parameters. Otherwise, if we do not have a strong belief, we set a uniform prior, which means it's equally likely to be true or false, and our model can still infer the truth from other factors. After drawing the $\theta_f$, we generate the truth label $l_f$ from a \textit{Bernoulli} distribution with parameter $\theta_f$:
\begin{equation}
l_f \sim Bernoulli(\theta_f).
\end{equation}

\vspace{0.05in}
\noindent \textbf{Pattern source reliability.} As aforementioned, a reliable pattern source is more likely to extract true facts with higher counts, and extract false facts with lower counts. Therefore, we choose average count of false/true as latent pattern reliable weight, it's represented as $\lambda_0^{{p}^{(*)}} $, $\lambda_1^{{p}^{(*)}} $ for pattern ${p}^{(*)}$. The Gamma distribution is utilized because it is the conjugate prior of Poisson distributions. Initially, these two parameters are generated from  \textit{Gamma} distribution with hyper-parameter $\{\mu_0,\kappa_0\}$/$\{ \mu_1,\kappa_1\}$, respectively. $\mu_0$ and $\mu_1$ represent the prior number of false/true fact the pattern extract, and $\kappa_0$ and $\kappa_1$ determine the prior sum of false/true fact count:
\begin{eqnarray}
\lambda_0^{{p}^{(*)}} \sim Gamma(\mu_0,\kappa_0); \\
\lambda_1^{{p}^{(*)}} \sim Gamma(\mu_1,\kappa_1)
\end{eqnarray}

\vspace{0.05in}
\noindent \textbf{Extraction observation.} For each extraction e $\in \mathcal{E}$, it is composed of $\{f,p^{(*)},o\}$.  $f_e$ denotes the temporal fact f belongs to e, $p^{(*)}$ denotes where it's extracted, $o_e$ stands for extraction $e$'s observation count. When the truth label of fact $f_e$ is false, $o_e$ is generated from \textit{Poisson} distribution with $p^{(*)}$'s false speaking side parameter  $\lambda_0^{{p}^{(*)}}$. While $f_e$ is true, $o_e$ is generated from Poisson Distribution with $p^{(*)}$'s true speaking side parameter $\lambda_1^{{p}^{(*)}}$:
\begin{equation}
\begin{aligned}
\begin{array}{lr}
o_e \sim Poisson(\lambda_0^{{p}^{(*)}})& \text{if} \ l_{f_e} = 0,\\
o_e \sim Poisson(\lambda_1^{{p}^{(*)}})& \text{if} \ l_{f_e} = 1.
\end{array}
\end{aligned}
\end{equation}

\vspace{0.05in}
\noindent \textbf{Constraints.} Finally, we draw the constraint variables. In temporal fact extraction, we define two variables $C_{1(e,t)-1v}$ and $C_{1v-1e}$. $C_{1(e,t)-1v}$ limits the number of truth on certain constraint key $\{e,t\}$. There are as many $C_{1(e,t)-1v}$ variables as unique $\{e,t\}$ keys:
\begin{equation}
\begin{aligned}
C_{e,t} = {\sum_{f} l_f}, \quad f \in \mathcal{F}_{e,t}, \\
\end{aligned}
\end{equation}
where $\mathcal{F}_{e,t}$ denotes a set of $f$ with same $\{e,t\}$. Each $C_{1(e,t)-1v}$ is generated by $\mathcal{F}_{e,t}$ set.

$C_{1v-1e}$ ilimits the truth of fact with same $\{v\}$:
\begin{equation}
\begin{aligned}
C_{v} = {\sum_{e \in E} l_{e,v}} \quad
\left\{
\begin{aligned}
\begin{array}{lr}
l_{e,v}  = 1,& \text{if} \quad  \exists l_f = 1, \\
& \quad f  \in \mathcal{F}_{e,v}; \\
l_{e,v}  =0,& \text{otherwise}.
\end{array}
\end{aligned}
\right.
\end{aligned}
\end{equation}
where $\mathcal{F}_{v}$ denotes set of fact with value $v$,  $\mathcal{F}_{e,v}$ stands for a set of temporal fact $f$ with same  $\{e,v\}$, $\mathcal{F}_{e,v} \in \mathcal{F}_{v} $. $l_{e,v}$ denoted the truth label of ${v,e}$. Each $\mathcal{C}_{1v-1e}$ is generated by $\mathcal{F}_{v}$. If there is true fact $f$  $\in \mathcal{F}_{e,v}$, then $l_{e,v}$ equals to one, otherwise, $l_{e,v}$ equal with zero.

\begin{table*}[t]
\centering
\caption{Our proposed model performs better than baseline methods on finding temporal facts.}
\begin{tabular}{|l|p{2cm}|p{2cm}||l|l|l|l|}
\hline
\multirow{3}{*}{\textbf{Method}}&
\multicolumn{2}{c|}{\textbf{Constraints}} & \multicolumn{4}{c|}{\textbf{Evaluation Setting}} \\
\cline{4-7}
& \multicolumn{2}{c|}{$e$: \texttt{Country}; $v$: \texttt{Person}; $t$: year}&\multicolumn{2}{c|}{On ($e$,$v$,$t$)} & \multicolumn{2}{c|}{On ($e$,$v$,[$t_{min}$,$t_{max}$])}\\
\cline{2-7}
& $\mathcal{C}_{1v-1e}$ & $\mathcal{C}_{1(e,t)-1v}$ & \textbf{AUC} & \textbf{F1} &\textbf{AUC} & \textbf{F1}\\
\hline
\textsc{TruthFinder} & \cmarkb & \xmark & 0.0006  & 0.0012 & 0.0006 & 0.0012 \\
\hline
\textsc{LTM}  & \xmark & \xmark & 0.1319  & 0.0199 & 0.2030  & 0.0218 \\
\hline
\textsc{LTM}  & \xmark & \cmarkb  & 0.0212  & 0.0505 & 0.0407 & 0.0793 \\
\hline
\textsc{TruePIE}  & \cmarkb & \xmark & 0.0587  & 0.1430 & 0.0587 & 0.1430 \\
\hline
\textsc{MajVote} & \xmark & \cmarkb & 0.3336  & 0.4318 & 0.4958 & 0.5927 \\
\hline
\textsc{TFWIN}  & \cmarkb & \cmarkb & 0.4746  & 0.6361 & 0.5523 & 0.6489 \\
\hline
\hline
Ours (\textsc{PGMCC}) & \xmark & \cmarkb  & 0.4840  & 0.6502 & 0.6006 & 0.7254 \\
\hline
Ours (\textsc{PGMCC}) & \cmarkb  & \cmarkb  & \textbf{0.4987}  & \textbf{0.6634} & \textbf{0.6075} & \textbf{0.7316} \\
\hline
\end{tabular}
\label{exp_result}
\end{table*}

\section{Experiments}
\label{sec:experiments}

\subsection{Dataset}

We focus on attribute \textit{country's president} and experiment on the same data set in the work of \cite{wang2019novel}. It has 9,876,086 news articles (4 billion words) published from 1994--2010. 
We have {57,472} patterns, {116,631} temporal fact tuples, and {1,326,164} extractions.
The dataset's ground truth was collected from Google and Wikipedia. It includes 3,175 true temporal facts of 130 countries.

\subsection{Experiment Settings}

\subsubsection{Competitive methods}
We compare our model with:\\
$\bullet$ \textsc{TruthFinder} \cite{yin2008truth}: It was a bootstrapping algorithm for structured data using $C_{1v-1e}$. \\
$\bullet$ \textsc{LTM}\cite{zhao2012probabilistic}: It was a probabilistic model, assuming that the truth about an object contains more than one value. We set ``object'' as \{entity, time\} and set value as the temporal fact's value. \\
$\bullet$ \textsc{TruePIE} \cite{qli2018truepie}: It was a bootstrapping method using $\mathcal{C}_{1(v)-1e}$ and estimating pattern reliability. \\
$\bullet$ \textsc{MajVote} \cite{goldman1995learning}: It used the weighted majority voting strategy and returned the most frequent temporal fact. \\
$\bullet$ \textsc{TFWIN} \cite{wang2019novel}: It was the state-of-the-art bootstrapping method for truth discovery on fact extraction. However, error propagation is serious in its iterative process.

\subsubsection{Evaluation settings}
All the methods can only find truth of temporal fact at one time point, e.g., (French, Jacques Chirac, 1995). However, due to the incompleteness of fact description in data, some time points of temporal facts could be missing. One way to improve the evaluation is to composite true temporal fact time points $\{e,v,t\}$ into temporal fact time period $\{e,v,[t_{min},t_{max}]\}$.
We evaluate the performance on both temporal fact time point $\{e,v,t\}$ and temporal fact time period $\{e,v,[t_{min},t_{max}]\}$. To evaluate on time period $\{e,v,[t_{min},t_{max}]\}$, we look at every single time points $(e,v,t)$ in the period ($t \in [t_{min},t_{max}]$).
 
\subsubsection{Evaluation metrics}
We evaluate all competitive methods using \textit{precision}, \textit{recall}, \textit{F1 score}, and \textit{AUC} (Area Under the Curve).
Precision is the the fraction of temporal fact truth among all the temporal fact that were labelled as true. Recall is the fraction of true temporal facts our approach finds among the ground truth temporal facts. F1 score is the harmonic mean of precision and recall. For all of the metrics, higher score indicates that the method performs better.

\subsection{Effectiveness}

The results are given in Table~\ref{exp_result}. Our proposed method \textsc{PGMCC} consistently outperforms all the baselines on finding (country, president, time)-facts (i.e., presidential terms).

\vspace{0.05in}
\noindent \textbf{\textsc{PGMCC} vs \textsc{LTM}:} \textsc{PGMCC} performs significantly better than \textsc{LTM} (+34.5\% AUC; +64.4\% F1) on evaluating time points, and performs better with (+40.45\% AUC; +71.2\% F1) on evaluating time periods. \textsc{LTM} was designed to solve structured truth finding like the bookseller example. So, there were many conflicts when applied to temporal fact extraction. \textsc{PGMCC} has multi-constraint as observable variables to alleviate the issue.

\begin{table*}[t]
\centering
\caption{Pattern's reliability for {country's presidency}.}
\begin{tabular}{|l|l|l|}
\hline
\textbf{Textual Pattern} $\mathbf{p}$ & $r_{p^{(post)}}$ & $r_{p^{(tag)}}$ \\ \hline \hline
president \texttt{Person} of \texttt{Country} & 0.920 & 0.870 \\
\texttt{Country}'s current president \texttt{Person}, & 0.978 & 0.250 \\
\texttt{Country}'s newly elected president , \texttt{Person} , & 0.970 & 0.030 \\
\texttt{Person}, now president of \texttt{Country}, & 0.750 & 0.110 \\
\hline
\texttt{Person}, who has ruled \texttt{Country} & 0.438 & 0.994 \\
\$\textsc{Country}'s former president \texttt{Person} & 0.113 & 0.994 \\
\hline
\texttt{Person}, who ruled \texttt{Country} & 0.607 & 0.758 \\
\texttt{Country} president \texttt{Person} signed  & 0.553 & 0.327 \\
\hline
\texttt{Country} premier \texttt{Person} & 0.012 & 0.010\\
\texttt{Country} foreign minister \texttt{Person} & 0 & 0 \\
\texttt{Country} golfer \texttt{Person} & 0 & 0 \\
\hline
\end{tabular}
\label{tab:casestudy}
\end{table*}

\vspace{0.05in}
\noindent \textbf{\textsc{PGMCC} vs \textsc{TFWIN}:} \textsc{PGMCC} performs better than \textsc{TFWIN} (+2.4\% AUC; +2.8\% F1) on evaluating time points, and performs better with (+5.2\% AUC; +8.3\% F1) on evaluating time periods. \textsc{TFWIN} started with seed patterns and defined constraints as a rule to eliminate conflicting tuples. However, the inference on conflicts was based on local information (i.e., the current pattern reliability estimation). During this process, error might propagate through iterations. \text{PGMCC} is a probabilistic graphical model that can avoid error propagation by modeling constraints as variables and inferring truth with the global data distributions.

\vspace{0.05in}
\noindent \textbf{\textsc{PGMCC} with different constraints:} See the last two rows in Table~\ref{exp_result}. For both \textsc{PGMCC} and \textsc{TFWIN} models, a complete constraint set, i.e., \{$\mathcal{C}_{1(v)-1e}$ and $\mathcal{C}_{1(e,t)-1v}$\}, gives the best performance. Partial constraint cannot fully identify conflicts or false tuples. $\mathcal{C}_{1(e,t)-1v}$ plays a significant role in extracting \textit{country's president}.

\subsection{Pattern Source Reliability Analysis}

Table~\ref{tab:casestudy} presented some pattern examples and their scores. Here are our observations. First, the pattern ``president \texttt{Person} of \texttt{Country}'' is the only pattern that shows high reliability on both types of time signals (above 0.85). Second, the textual patterns that describe the current presidency are likely to have higher reliability on \textit{text gen. time} than \textit{temporal tag}, because the presidency was likely to be in the same time as the document was generated. These patterns usually have words such as ``current'', ``newly'', and ``now''. Third, the textual patterns that describe the past presidency are likely to have higher reliability on ``tag'' than ``post'', because the presidency was likely to be in the same time as the event (described in the sentence) happened but before the time of the document being generated. These patterns usually have words such as ``have governed'', ``have ruled'', ``former'', and ``formerly''.

\section{Related Work}
\label{sec:related}

In this section, we review two relevant fields to our work, {temporal fact extraction} and {truth discovery}.

\subsection{Truth Discovery}
In big data era, the issue of ``Veracity'' on resolving conflicts among multi-source information is quite serious \cite{berti2015data,vydiswaran2011content,waguih2014truth,dong2009integrating,galland2010corroborating,xiao2016towards,yin2011semi}. Truth discovery methods find trustworthy information from conflicting multi-source \cite{xiao2015believe,li2015discovery}. Several truth discovery methods have been proposed for various scenarios, and they have been successfully applied in diverse application domains. A few truth discovery methods are probabilistic model. \textsc{LTM} solved the ``\texttt{Book}'s \texttt{author} list problem'' and modeled its source in two-fold quality \cite{zhao2012bayesian}. \textsc{GTM} solved the task of finding true numeric value of ``New York City's population'' \cite{zhao2012probabilistic}. \textsc{TextTruth} found the true answer for a question from multi users \cite{zhang2018texttruth}. 

\subsection{Temporal Fact Extraction}
Temporal fact extraction is to extract (entity, attribute name, attribute value)-tuples along with their time conditions from text corpora \cite{sil2014towards,hoang2016unified,chekol2017scaling,zhang2018taxogen,shang2018automated,zeng2019faceted,jiang2019role}. Textual patterns have been proposed to extract structured data from unstructured text data in an unsupervised way, such as E-A patterns \cite{gupta2014biperpedia}, parsing patterns \cite{nakashole2012patty}, and meta patterns \cite{jiang2017metapad}. However, patterns are of different reliability and extractions are sometimes conflicting. In order to get reliable temporal fact, we addressed this problem using truth discovery.

\section{Limitations and Future Work}

Though the proposed approach show effectiveness in experiments, it and/or the study has several limitations. First of all, because collecting \emph{temporal} factual truth for a variety of attributes is very expensive, in this study, we only studied a single relation type. In future work, we will apply the approach to other types of temporal-facts if correct constraints can be defined, such as sports team's players and spouse relationship. Second, though the patterns were generated by automated mining technologies such as Meta Patterns \cite{jiang2017metapad} (in other words, they are not hand-crafted), the pattern mining as a preprocessing step is needed. The approach is not end-to-end.

\section{Conclusions}
\label{sec:conclusions}

In this work, we proposed a probabilistic graphical model for inferring true facts and pattern reliability. It had two novel designs for temporal facts: (1) it modeled pattern reliability on temporal tag in text and text generation time; (2) it modeled commonsense constraints as observable variables. Experimental results demonstrated that our model outperformed existing methods.

\section*{Acknowledgements}
This work was supported by NSF Grant IIS-1849816 and CCF-1901059.

\balance
\bibliography{paper}
\bibliographystyle{acl_natbib}

\end{document}